\pdfoutput=1

\documentclass[11pt]{article}

\usepackage[preprint]{acl}

\usepackage{times}
\usepackage{latexsym}
\usepackage{xfrac}

\usepackage[T1]{fontenc}

\usepackage[utf8]{inputenc}

\usepackage{microtype}

\usepackage{inconsolata}

\usepackage{graphicx}

\usepackage{multirow}

\usepackage{xspace}
\usepackage{paralist}
\usepackage{amsmath,amsthm,mathtools,amssymb}
\usepackage{booktabs}
\usepackage{todonotes}

\usepackage{color-edits}
\addauthor{ms}{cyan}

\newtheorem{proposition}{Proposition}
\theoremstyle{definition}
\newtheorem{assumption}{Assumption}

\newcommand{\methodname}{SYNAPSE-G\xspace}

\usepackage{algorithm}
\usepackage{algorithmic}
\usepackage{graphicx}
\usepackage{wrapfig}
\usepackage{subcaption}
\usepackage{tikz}
\usetikzlibrary{shapes.geometric, arrows.meta, positioning, shadows} 

\title{SYNAPSE-G: Bridging Large Language Models and Graph Learning for Rare Event Classification}

\author{Sasan Tavakkol \\
  Google Research \\ \And
  Lin Chen \\
  Google Research \\ \And
  Max Springer \\
  University of Maryland \\
  Google Research \\ \And
  Abigail Schantz \\
  Google Research \\ \AND
  Blaž Bratanič \\
  Google Research \\ \And
  Vincent Cohen-Addad \\
  Google Research \\ \And
  MohammadHossein Bateni \\
  Google Research \\
  }

\begin{document}
\maketitle
\begin{abstract}
    Scarcity of labeled data, especially for rare events, hinders training effective machine learning models. 
This paper proposes \methodname (\underline{Syn}thetic \underline{A}ugmentation for \underline{P}ositive \underline{S}ampling via \underline{E}xpansion on \underline{G}raphs), a novel pipeline leveraging Large Language Models (LLMs) to generate synthetic training data for rare event classification, addressing the cold-start problem. 
This synthetic data serve as seeds for semi-supervised label propagation on a similarity graph constructed between the seeds and a large unlabeled dataset. 
This identifies candidate positive examples, subsequently labeled by an oracle (human or LLM). 
The expanded dataset then trains/fine-tunes a classifier. 
We theoretically analyze how the quality (validity and diversity) of the synthetic data impacts the precision and recall of our method. 
Experiments on the imbalanced SST2 and MHS datasets demonstrate \methodname's effectiveness in finding positive labels, outperforming baselines including nearest neighbor search. 
\end{abstract}

\section{Introduction} \label{sec:introduction}
The rapid emergence of new trends on social media and the internet, such as misinformation~\cite{suarez2021prevalence}, fraud~\cite{herland2019effects}, and hate speech~\cite{mathew2019spread,siegel2020online}, necessitates the development of effective classifiers for timely detection and mitigation~\cite{banks2010regulating}. However, the dynamic nature and novelty of these trends often results in scarcity of labeled data for training supervised models~\cite{shyalika2024comprehensive}. This research tackles this ``cold-start'' problem by introducing a pipeline that leverages LLMs and semi-supervised learning for collecting labeled data for training robust classifiers. Our method, \methodname, offers a practical and generalizable solution for addressing emerging online threats.

\methodname augments real labeled data with LLM-generated synthetic data in three stages: (1) \textbf{Data Generation:} An LLM generates rare event examples (e.g., hate speech), creating a seed set. (2) \textbf{Label Propagation:} This seed set is used in semi-supervised label propagation, expanding the labeled set by connecting seeds to similar unlabeled instances on a similarity graph. (3) \textbf{LLM-Based Refinement (Optional):} An LLM (or human) rater can refine propagated labels, mitigating errors. 
The augmented dataset then trains/fine-tunes a classifier.
In industrial practice, \methodname has been deployed to evaluate content against a substantial number of abuse policies in a single request. 
Given the quantity and specificity of these policies, labeled data for initial training was unavailable or extremely rare for many of them, which led to the formulation and successful implementation of our method.

This paper proceeds to make three major contributions:
(1) We propose \methodname to combine LLM-generated synthetic data with graph-based semi-supervised learning for rare event classification.
(2) We theoretically analyze how the validity and diversity of synthetic data affect the precision and recall of our label propagation.
(3) We empirically demonstrate the effectiveness of our method on public datasets, demonstrating strong gains over competitive baselines.



\section{Related Work} \label{sec:related_work}

\paragraph{Retrieval.}
Retrieval methods aim to identify relevant items from a large dataset based on a given query.
Traditional approaches rely on lexical matching techniques such as BM25~\cite{robertson1994some}, which score documents based on term frequency and inverse document frequency (TF-IDF).
While effective for keyword-based search, these methods fail to capture semantic meaning limiting their performance on more complex retrieval tasks.
To overcome these limitations, dense retriever methods have emerged, leveraging neural embeddings to map both queries and documents into a shared vector space where relevance can be measured with standard similarity metrics~\cite{karpukhin2020dense,lee2019latent,xiong2020approximate,izacard2021unsupervised}.
Closely related to our work is that Hypothetical Document Embeddings (HyDE)~\cite{gao2022precise}, which bypasses the need for relevance labels by generating a synthetic document which is then used to retrieve similar (real) documents from a dataset, allowing for effective search without fine-tuning or task-specific supervision.

\paragraph{Diversity Sampling.}
Collecting high-quality human-labeled data is often costly or infeasible, particularly due to privacy concerns and annotation overhead \cite{kurakin2023harnessing}. 
Moreover, recent studies have highlighted that human-generated data can exhibit systematic biases or inconsistencies, making it suboptimal for model training across a range of tasks \cite{gilardi2023chatgpt,hoskinghuman,singhbeyond}. 
To address these limitations, emerging research has focused on synthetic data generation as a means of more diversely sampling the training space \cite{gandhi2024better,liu2024best}. 
In our setting, the target data\textemdash{}rare positive examples\textemdash{}are unlikely to be captured through random sampling alone. 
We therefore leverage synthetic data to steer our graph-theoretic exploration toward this small, informative subset, enabling more effective identification and labeling.

\paragraph{Positive Mining.}
Positive (or negative) mining seeks to identify instances that are likely to belong to a target (our non-target) class, often used to help refine decision boundaries in the data space especially when labeled data is scarce or imbalanced~\cite{qu2020rocketqa}.
Traditional approaches, such as hard negative mining in contrastive learning~\cite{xiong2020approximate}, select negatives that are close to the decision boundary to improve model generalization~\cite{hofstatter2021efficiently}.
In our setting, positive mining is at the core of detecting rare events within a large unlabeled dataset.

\paragraph{Label Propagation on Graphs.}
Our work falls in the domain of ``label propagation'', a fundamental approach in semi-supervised learning that leverages the structure of data to infer labels for unlabeled instances.
This method assumes that similar points should have similar labels, enforcing smoothness in the label distribution~\cite{bengio2006label}.
However, label propagation relies on the presence of an initial labeled dataset and assumes that the underlying graph structure accurately captures class boundaries~\cite{talukdar2014scaling,ravi2016large,baluja2008video}.

In contrast, our approach addresses a fundamentally different problem: identifying rare, positive, instances within a large \emph{unlabeled} dataset without an existing set of labeled examples.
Rather than relying on label propagation from known labels, we generate synthetic instances for the rare event and use their embedding to identify similar real instances.
This removes the need for model training or iterative graph-based updates and makes our approach particularly suitable for applications where positive instances are extremely rare and must be identified without prior ground truth labels.

\section{Preliminaries \& Problem Definition} \label{sec:problem_formulation}
This work tackles the problem of binary classification for domains where a ``rare event" class is significantly underrepresented.
Formally, let \(\mathcal{D}\) denote the data domain. Each observation is represented by a feature vector \(x\), and the data distribution is denoted by \(\Pr(x, y)\), where \(y \in \{0, 1\}\) is the class label (\(y=1\) is the rare event).
Deviating from supervised learning, we confront a \textbf{complete absence of labeled data} initially. 
We operate within an active learning framework, tailored for iterative label acquisition. 
An algorithm begins with unlabeled dataset \(\mathcal{D}_U = \{x_j\}_{j=1}^{n_U}\) and, iteratively, a batch of data is selected according to some strategic mechanism wherein an oracle is queried to label the data.
Subsequently, the learning algorithm is updated according to the new labels.


We seek to maximize the cumulative precision and recall across all queried batches. 
Formally, at each step \(i\), let \(P_i\) and \(R_i\) represent the precision and recall, respectively, calculated over the union of all labeled sets, $\mathcal{L}_j$ acquired up to that point: \(\bigcup_{j=1}^{i} \mathcal{L}_j\). 
The algorithm aims to maximize both \(P_i\) and \(R_i\) for all \(i \in \{1, ..., T\}\). 
This reflects the goal of efficiently identifying as many rare events as possible with minimal false positives. The core challenges are the \textbf{cold start} (no initial labeled data) and the \textbf{severe class imbalance}.

\section{Methodology} \label{sec:method}
\begin{figure*}[htbp]
\centering
\resizebox{0.85\linewidth}{!}{
\begin{tikzpicture}[
    node distance=1.5cm,
    font=\LARGE,
    block/.style={rectangle, draw, text width=8em, align=center, rounded corners, minimum height=3em, drop shadow,fill=white},
    db/.style={cylinder, shape border rotate=90, draw, aspect=0.5, text width=8em, align=center, drop shadow},
    arrow/.style={->, >=stealth, line width=1.5pt, >={Stealth[length=5mm]}}
]

    \node (concept) [block] {Concept Definition};
    \node (gen_sd) [block, right=of concept] {Generate Synthetic Data (SD)};
    \node (ada_samp) [block, right=of gen_sd] {Adaptive Sampling
};
    \node (embed_sd) [block, right=of ada_samp] {Embed Synthetic Data};
    \node (pos_seeds) [block, right=of embed_sd] {Create Set of  Seeds};
    \node (prop_labels) [block, right=of pos_seeds] {Propagate Labels};
    \node (candidates) [block, right=of prop_labels] {Select Top Candidates};
    \node (rate) [block, right=of candidates] {Rate w/ Oracle};
    \node (labeled_data) [db, right=of rate, fill=yellow!20] {Labeled Hybrid Data};

    \node (unlabeled_data) [db, below=1.5cm of ada_samp, fill=gray!20] {Unlabeled Real Data};
    \node (embed_rd) [block, right=of unlabeled_data] {Embed Real Data};
    \node (bipartite) [block, right=of embed_rd] {Build a  Similarity Graph};

    \node (model_training) [block, dashed, below=of labeled_data] {Model Training};

    \draw [arrow] (concept) -- (gen_sd);
    \draw [arrow] (gen_sd) -- (ada_samp);
    \draw [arrow] (ada_samp) -- (embed_sd);
    \draw [arrow] (embed_sd) -- (pos_seeds);
    \draw [arrow] (pos_seeds) -- (prop_labels);
    \draw [arrow] (prop_labels) -- (candidates);
    \draw [arrow] (candidates) -- (rate);
    \draw [arrow] (rate) -- (labeled_data);

    \draw [arrow] (unlabeled_data) -- (embed_rd);
    \draw [arrow] (embed_rd) -- (bipartite);

    \draw [arrow] (pos_seeds) -- (bipartite);
    \draw [arrow] (bipartite) -- (prop_labels);
    \draw [arrow] (labeled_data) -- (model_training);

    \draw [arrow] (rate) -- ++(0,2.0) -| (pos_seeds); 
    \draw [arrow] (gen_sd) -- ++(0,3.5) -| (labeled_data.north); 

\end{tikzpicture}
}
\caption{\small Overview of the \methodname pipeline. The pipeline integrates synthetic data generation (top branch) with real data processing (bottom branch).  LLM-generated synthetic data, after adaptive sampling and embedding jointly with unlabeled data, forms a set of positive seeds.  
A similarity graph connects seeds and real data, enabling label propagation to identify top candidates. 
An oracle rates these candidates, creating a labeled hybrid dataset for model training in an iterative feedback loop.}
\label{fig:pipeline}
\end{figure*}
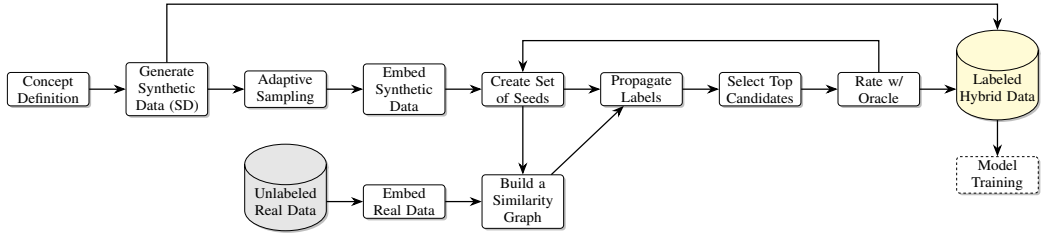

Our method addresses rare event classification with limited labeled data by generating and leveraging synthetic data.  
The core is a three-stage pipeline integrating LLMs with semi-supervised learning to augment a small (or non-existent) initial set of labeled real data. 
Figure \ref{fig:pipeline} provides an overview.

\subsection{Synthetic Data Generation} \label{sec:synthetic_data_generation}

To address the cold start problem (absence of initial labeled data), we use an LLM to generate an initial seed set of labeled data, $\mathcal{D}_S$, to bootstrap the learning process.  
In practice, one should use carefully crafted prompts to guide the LLM towards generating examples representative of the rare event class. 
However, we omit this step here as it is outside the scope of this research, and instead focus on selecting the best data points from a pool of synthetically generated data. 
We will compare two selection methods in the experiments: random sampling and the Adaptive Coverage Sampling (ACS) approach of~\cite{tavakkol2025less} which selects $k$ points that collectively cover a $c$-portion of the dataset, maximizing diversity within the synthetic data.

\subsection{Label Propagation to Unlabeled Data}

We subsequently expand the labeled dataset by propagating labels from the synthetic seed data ($\mathcal{D}_S$) to the unlabeled data ($\mathcal{D}_U$) via semi-supervised learning.
We assume access to a large corpus of unlabeled data, $\mathcal{D}_U$, representative of the target domain and containing both rare event and non-event instances. 
Both synthetic and unlabeled data are transformed into numerical embeddings (e.g., using BERT~\citep{devlin2018bert} or Gecko \citep{lee2024gecko}) so that semantically similar data points are close in the embedding space.
We propose two semi-supervised approaches which we denote as \emph{Iterative Bipartite Graph (IBG)} and \emph{Label Propagation (LP)}.

\paragraph{Iterative Bipartite Graph (IBG)}
IBG iteratively refines a bipartite graph between known positives ($V_P$, initially $\mathcal{D}_S$) and unlabeled data ($V_U$). 
Edges are created based on cosine similarity exceeding a threshold, then pruned to retain only the top $d_{max}$ connections per node in $V_P$. 
Connected $V_U$ nodes are queried for labels. 
New positives are added to $V_P$, and the process repeats. 
The pseudocode for this process is provided in Appendix~\ref{apx:pseudo}.

\paragraph{Label Propagation (LP)}
This approach utilizes a more global graph structure and a standard Label Propagation algorithm \citep{zhu2002learning,ravi2016large}. 
The specific algorithm is detailed as pseudocode in Appendix~\ref{apx:pseudo}.
In brief, a similarity graph is constructed over the entire real dataset and the initial synthetic seed data. 
The known labels (positive and negative) are then propagated through this graph by a learned weight to each real data point (its likelihood of being positive), and the top $K$ points are selected as candidates for labeling. 
$K$ is dynamically adjusted each iteration: $K = K_0 / p_{\text{prev}}$, where $p_{\text{prev}}$ is the precision from the previous iteration, aiming to find approximately $K_0$ new positives per round.

We note that, in large-scale settings, the computational cost of creating and optimizing the similarity graph structure (for ACS or LP) can become prohibitive with a time complexity of $O(n^2)$, if implemented via a naive pairwise computation of similarities.
However, we significantly speeded up such computations with methods such as Locality Sensitive Hashing (LHS) or hop-spanner methods~\cite{carey2022stars,epasto2021clustering,halcrow2020grale} for scalable deployment of SYNAPSE-G. In addition to the graph construction step, scalable variants of other graph techniques can also be deployed that rely on optimizing for a small random subset of the data and utilizing the optimized parameters for the full graph~\cite{tavakkol2025less}.

\section{Theoretical Analysis} \label{sec:theory}
To understand how prompt quality (validity and diversity of synthesized data) impacts algorithm performance, we analyze a simplified, single-iteration version of our algorithm. 
We model data and relationships using an undirected, simple, $d$-regular graph $G=(V,E)$. 
Each node $v \in V$ is a data point with a binary label $y(v) \in \{0, 1\}$ (1: positive, 0: negative). 
Let $S \subseteq V$ represent the LLM-generated synthesized data. 
The algorithm queries $S$ and then the neighbors of positive seeds, $N(S_+)$.
In proving theoretical results on our algorithms expected guarantees, we first invoke a few careful assumptions on the input which hold in practice.

\begin{assumption}[Diversity of Synthesized Data] \label{assump:diversity}
$S$ exhibits two properties. (1) \emph{Independence}: $S$ forms an independent set on the graph. (2) \emph{Limited Overlap}: No vertex in $V$ is adjacent to more than two vertices in $S$.
\end{assumption}

We define a partition of $S$ into $S_+ = \{v \in S \mid y(v) = 1\}$ and $S_- = \{v \in S \mid y(v) = 0\}$. 
Thus, we define $p = \frac{|S_+|}{|S|}$ as the proportion of positive examples (validity).
Furthermore, if $u \in V$ is adjacent to exactly $n$ positive vertices in $S$, then denote the probability that $u$ is labeled by $q_n$ 
for $n = 1, 2$. 
Assume $0 < q_1 < q_2 < 2q_1$.\footnote{A positive example independently assigns a positive label to a neighbor with probability $q_1$. Thus, $q_2 = 1 - (1-q_1)^2$.}
We now obtain the following proposition with the proof deferred to Appendix~\ref{apx:proofs} due to space constraints.

\begin{proposition} \label{prop:precision-recall}
Let \( Q \coloneqq S \cup N(S_+) \) be the set of queried vertices, and let \( P \) be the number of positives in \( Q \). Then $\mathbb{E}\left[\frac{P}{|Q|} \mid S\right]$ and $\mathbb{E}\left[\frac{P}{|V|} \mid S\right]$ are respectively:
\footnotesize{
\begin{gather}
(2q_1 - q_2) + 
\frac{1 + q_2(d + \sfrac{1}{p}) - q_1(d + \sfrac{2}{p})}{\sfrac{1 - p}{p} + h(S_+)} \\
\frac{(1 - 2q_1 + q_2) + (q_2 - q_1)d + (2q_1 - q_2)h(S_+)}{\sfrac{|V|}{p|S|}}
\end{gather}}
\end{proposition}

This result explores how two key dimensions of synthesized data quality – \emph{validity}, $p$, and \emph{diversity}, $h(S_+)$ – impact precision and recall. 
Intuitively, this first equality proves that recall increases with both $p$ (higher probability of synthesized seeds being truly positive) and $h(S_+)$ (greater diversity, allowing exploration of more examples). 
The relationship between precision and diversity, however, is more nuanced. 
Precision always increases with $p$, but the impact of diversity on precision depends on the magnitude of $p$ relative to a threshold determined by $q_1$, $q_2$, and $d$.  
This threshold, $\frac{2q_1 - q_2}{1 + (q_2 - q_1)d}$, is increasing in $q_1$ and decreasing in $q_2$.
In segregating the results based on this thresholding, we can conclude the following important two facts:

\noindent (1) When the validity of the synthesized positives is sufficiently high $\left(p > \frac{2q_1 - q_2}{1 + (q_2 - q_1)d}\right)$, precision \emph{decreases} with increasing diversity. 
Intuitively, if neighbors of single positive seeds are unlikely to be positive (low $q_1$), then maximizing precision requires focusing on regions with \emph{overlapping} neighborhoods (lower $h(S_+)$), increasing the chance of finding nodes adjacent to \emph{multiple} positive seeds (higher $q_2$).

\noindent (2) When the validity is low $\left( p < \frac{2q_1 - q_2}{1 + (q_2 - q_1)d} \right)$, precision \emph{increases} with diversity. 
In this case, even nodes adjacent to a \emph{single} positive seed are sufficiently likely to be positive, making greater coverage (higher $h(S_+)$) beneficial for precision.

This analysis reveals a crucial interplay between the validity and diversity of synthesized data and their combined effect on precision, offering valuable insights for designing effective prompt engineering and data selection strategies.

\section{Experimental Results} \label{sec:experiments}
We here validate \methodname on two representative datasets: the Stanford Sentiment Treebank 2 (SST2~\cite{socher2013recursive}) and Measuring Hate Speech (MHS~\cite{kennedy2020constructing, sachdeva2022measuring}). 
Table \ref{tab:dataset_stats_combined} summarizes dataset statistics.

\subsection{SST2 Dataset}
The SST2 dataset is a standard sentiment analysis benchmark comprised of movie review sentences with positive/negative labels. We augment this data with a public synthetic SST2 dataset generated using GPT \citep{ding2023gpt}. 

To simulate a rare event, we create a class-imbalanced SST2 training set, keeping all negative examples and subsampling positive examples to 10\% of the modified set. We select 100 positive synthetic examples as seeds.


\begin{table}[t!]
\centering
\resizebox{0.95\columnwidth}{!}{%
\begin{tabular}{lccc|ccc}
\toprule
\multirow{2}{*}{Dataset} & \multicolumn{3}{c|}{SST2} & \multicolumn{3}{c}{MHS} \\
                         & All   & Pos.  & Neg.      & All   & Pos.  & Neg. \\
\midrule
Train (Original / MHS)              & 67349 & 37569 & 29780 & 39565 & 2598  & 36967 \\
Synthetic (All)                     & 5000  & 2488  & 2512  & 1000  & 19    & 981   \\
Train (Imbalanced / --)            & 33088 & 3308  & 29780 & --    & --    & --    \\
Synthetic Seeds (Positive)         & 100   & 100   & 0     & 19    & 19    & 0     \\
\bottomrule
\end{tabular}
}
\caption{Dataset statistics for SST2 and MHS.}
\label{tab:dataset_stats_combined}
\end{table}
We focus on \emph{single-shot} (one iteration) and \emph{iterative} evaluations, using the imbalanced SST2 ``train'' split. 
The 100 positive seeds are selected from the synthetic dataset (randomly or via ACS~\cite{tavakkol2025less} with coverage $c=0.5$).  We construct a bipartite graph connecting seed and training examples, using cosine similarity of pre-trained Gecko embeddings~\citep{lee2024gecko} for edge creation.

\paragraph{Baselines}

We compare our approach against the following methods. \textit{Random Selection (Theoretical):} expected values, calculated analytically. 
\textit{Random Seeds + Bipartite Graph:} 100 random positive seeds; bipartite graph constructed as above and connected real data points are labeled. 
\textit{ACS Seeds + Bipartite Graph:} identical to (2), but using ACS for seed selection~\cite{tavakkol2025less}. 
\textit{Label Propagation:} similarity graph on the \emph{entire} real dataset + 100 initial ACS seeds. 
Labels are propagated via normalized adjacency (see Appendix~\ref{apx:pseudo} for full details).
The $K$ points of highest weights are selected as candidates, where $K = 100 / p_{\text{prev}}$ ($p_{\text{prev}}$: previous iteration's precision).

\paragraph{Results and Discussion}
We evaluate performance using precision-recall curves, recall v. query ratio, precision v. query ratio, and F1 score v. query ratio, analyzing the impact of similarity threshold and maximum degree ($d_{max}$)\footnote{These figures are enlarged in the appendix for maximal clarity.}.

\begin{figure}[t]
\centering
\begin{subfigure}[t]{0.22\textwidth}
  \centering
  \includegraphics[width=\linewidth]{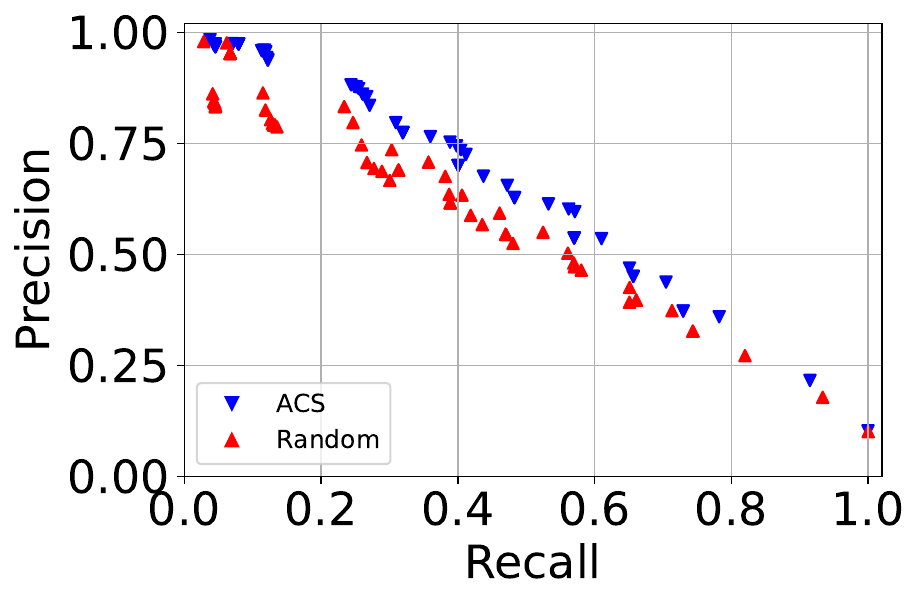}
  \caption{\small Precision v. Recall}
  \label{fig:image1}
\end{subfigure}\hfill
\begin{subfigure}[t]{0.22\textwidth}
  \centering
  \includegraphics[width=\linewidth]{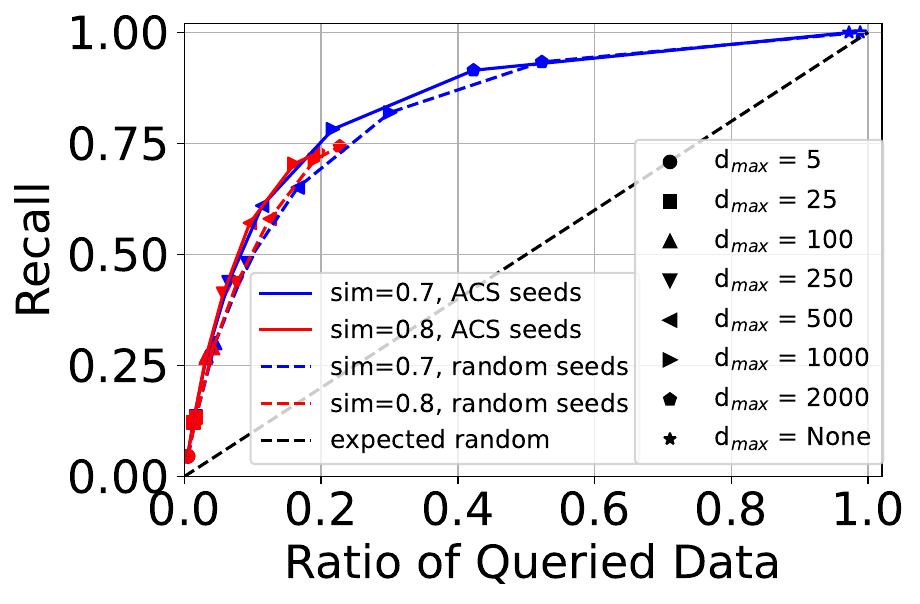}
  \caption{\small Recall v. Query Ratio}
  \label{fig:image2}
\end{subfigure}\hfill
\begin{subfigure}[t]{0.22\textwidth}
  \centering
  \includegraphics[width=\linewidth]{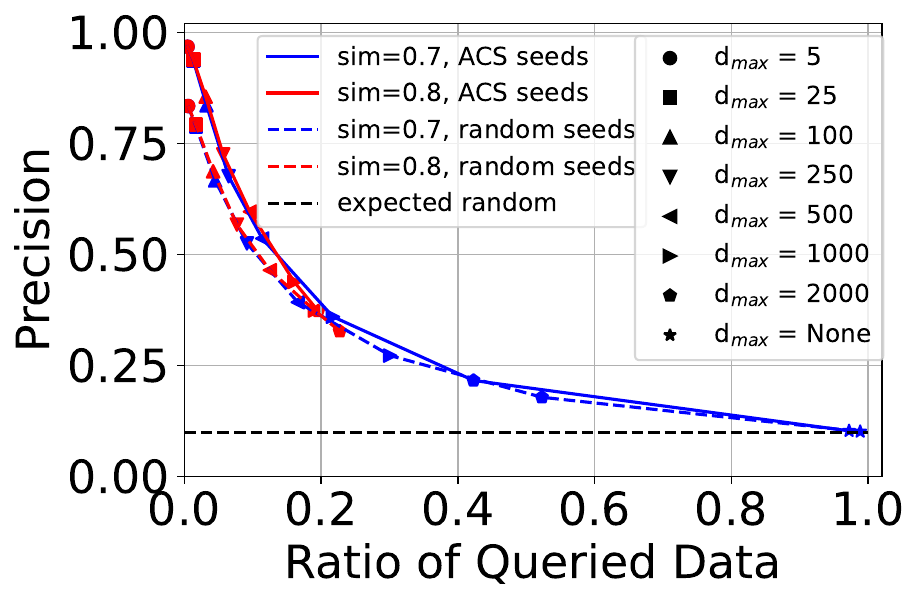}
  \caption{\small Precision v. Query Ratio}
  \label{fig:image3}
\end{subfigure}\hfill
\begin{subfigure}[t]{0.22\textwidth}
    \centering
    \includegraphics[width=\linewidth]{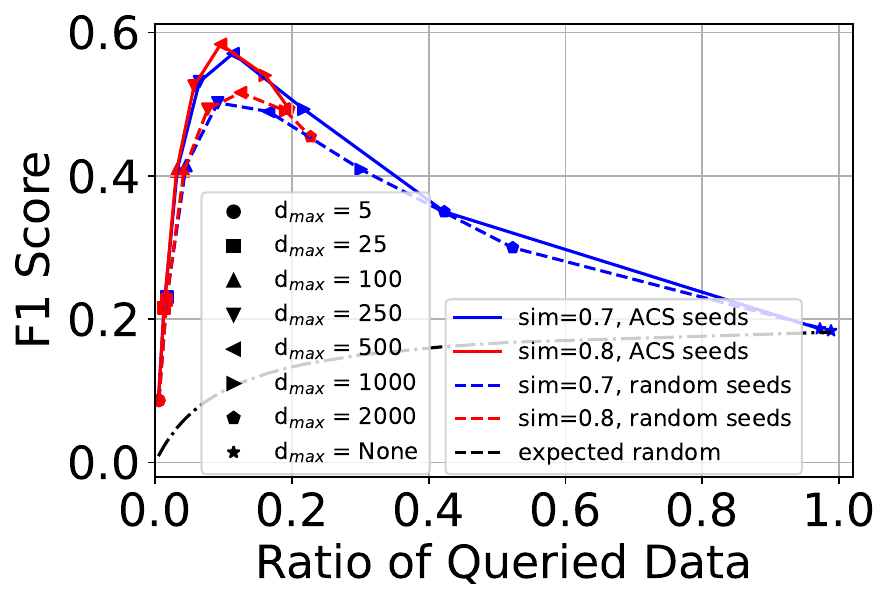}
      \caption{\small F1 Score v. Query Ratio}
    \label{fig:image4}
\end{subfigure}

\caption{\small Results for imbalanced SST2 dataset.
 }
\label{fig:combined_results}
\end{figure}

Figure \ref{fig:image1} shows that ACS seed selection consistently achieves higher precision for any given recall compared to random seed selection.  This highlights the benefit of diverse and representative seed sets.
Figure~\ref{fig:image2} plots the recall versus query ratio. 
The dashed line depicts expected recall from random performance. 
All methods here achieve a recall of 1.0 when querying all data, but for lower queries ACS seeds consistently achieve higher recall than random seeds. 
We note that a higher $d_{max}$ allows more connections in the bipartite graph, generally increasing recall, but with diminishing returns. 
A higher similarity threshold (0.8) generally results in better performance but limits reachability and consequently the recall.
Figure~\ref{fig:image3} show the precision versus query ratio, with the dashed line representing the base positive rate (10\%). 
Both graph-based methods (ACS and random seeds) achieve significantly higher precision than random selection, particularly at low query ratios and
ACS further outperforms random. 
Higher similarity thresholds and increasing $d_{max}$ (up to a point) generally improve precision.
Lastly, Figure \ref{fig:image4} shows the F1 score, with the black line representing expected random performance. 
Balancing precision and recall, F1 initially increases with the query ratio, peaks, and then decreases. 
ACS-selected seeds generally outperform random seeds. 
A higher similarity threshold leads to higher peak F1 scores. 
Increasing $d_{max}$ initially improves the F1 score, with diminishing returns and potential slight performance decreases at very high $d_{max}$ and high query ratios. 
The best F1 scores are below 0.6, highlighting the difficulty of the rare event detection problem.


\begin{figure}[t]
    \centering
\includegraphics[width=0.75\linewidth]{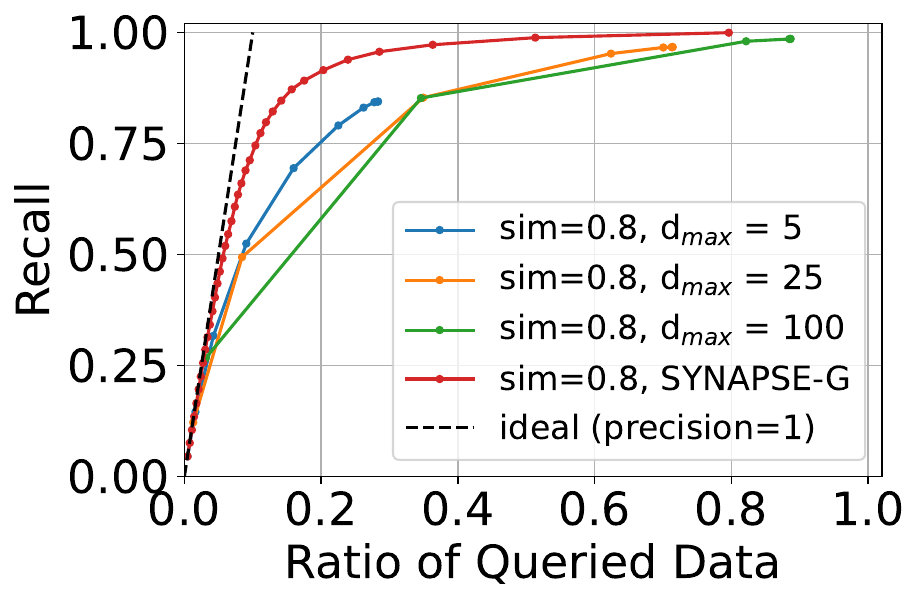}
\caption{\small Results for \emph{iterative} rare event detection on the imbalanced SST2 dataset. ``Ideal'' is perfect precision.} 
\label{fig:image5}
\end{figure}

Figure \ref{fig:image5} compares two iterative strategies, Iterative Bipartite Graph (IBG) and Label Propagation (LP), within an ``ideal'' scenario (precision = 1). 
LP significantly outperforms IBG, achieving much higher recall for a given query ratio. 
Notably, IBG plateaus quickly while LP leverages the full graph structure and both positive/negative labels.

\subsection{MHS Dataset}
To further evaluate SYNAPSE-G on a naturally occurring rare event task with real-world relevance, we utilized the MHS dataset \citep{kennedy2020constructing, sachdeva2022measuring}. 
This dataset contains 39,565 comments with annotations from 7,912 annotators (135,556 total rows). 

For labeling, the MHS dataset is more complex and relies on further preprocessing to define a specific rare event. 
We define our rare event as targeting transgender individuals.
This resulted in 2,598 comments (6.5\%) being labeled positive, representing an organically imbalanced dataset relevant to real-world challenges (see Appendix~\ref{apx:mhs-data} for a more careful breakdown of the rare-event designation). 
For the cold-start scenario, we used 1,000 LLM (Llama-2) generated comments related to the dataset's topics, sourced from \citep{casula2024delving}. 
Within this synthetic set, only 19 comments were identified with our target transgender label to serve as the initial positive seeds for SYNAPSE-G to identify real instances in the unlabeled data pool. 


\paragraph{Baselines}
We further design a practical baseline, ``LR-Baseline'', which adopts an iterative active learning approach using a simple classifier.
We establish the baseline using a logistic regression model 
trained on an initial set comprising 19 positive synthetic seeds augmented with 19 randomly sampled known negative instances from the dataset, all embedded with Gecko.
The iterative refinement process then proceeds as follows: in each iteration, a subset of unlabeled data points, constrained by an inference budget ($B$) is selected. 
The current logistic regression model predicts positivity probabilities for this subset, and the $K$ candidates with highest predicted probabilities are chosen for labeling via an oracle ($K = \sfrac{K_0}{p_{\text{prev}}}$ with $K_0 = 100$). 
Newly labeled instances are then incorporated into the training set, and the logistic regression model is retrained.

\begin{figure}[t]
    \centering
\includegraphics[width=0.75\linewidth]{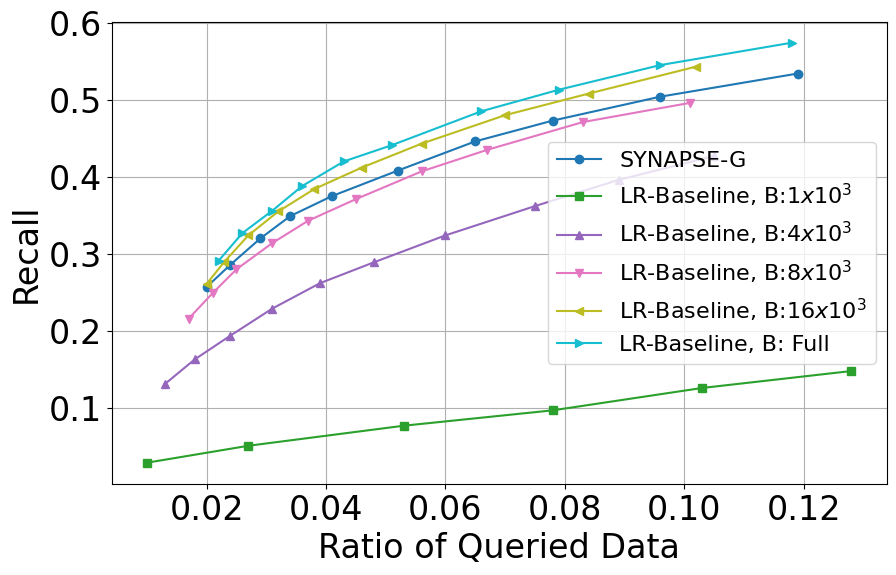}
\caption{\small Results for \emph{iterative} rare event detection on the imbalanced MHS dataset}
\label{fig:mhs}
\end{figure}

\paragraph{Results and Discussion}
Figure \ref{fig:mhs} summarizes SYNAPSE-G's recall performance 
compared against LR-Baseline's recall for inference budgets $\left(B = [1, 4, 8, 16] \times 10^3\right)$ and without a budget. Crucially, both SYNAPSE-G and LR-Baseline leverage the same Gecko embedding space, ensuring a fair comparison in terms of input features. 
It is important to note, however, that the LR-Baseline was initialized with access to 19 known true negative labels in addition to the 19 synthetic positive seeds, providing the LR-Baseline with an information advantage compared to SYNAPSE-G's strict cold-start setting, which assumes access only to synthetic positives and unlabeled data. 
Despite this, Figure~\ref{fig:mhs} reveals a clear trade-off between computational budget and recall. 
Crucially, \methodname remains highly efficient at discovering rare positive instances, achieving substantial recall with minimal labeling effort. 
For example, by labeling only 2.4\% of the data, SYNAPSE-G successfully identifies 28.6\% of the true positive comments. 
Increasing the labeling budget to just 5\% allows SYNAPSE-G to retrieve 40.8\% of the positives. 
In contrast, LR-Baseline requires a large inference budget ($B=8000$, inferring on ~20\% of the data per round) to reach similar recalls. 
Only with very large or unlimited budgets (thus, significant compute cost) does LR-Baseline outperform in terms of recall, highlight \methodname 's practical advantages.
While the current dataset exhibits moderate imbalance (6.5\% positive), real-world scenarios often present much more severe challenges (e.g., identifying a few thousand target posts among billions). 
By leveraging the graph structure to focus exploration around known positive seeds (synthetic or newly discovered real ones) and their neighbors, our method remains highly scalable and practical for discovering truly rare events in massive datasets.

\section{Conclusion}\label{sec:conclusion}
\methodname offers a compelling approach to the challenging task of rare event classification. Our theoretical underpinnings illuminate the critical balance between the fidelity and diversity of the synthesized data, providing insights into the method's efficacy while empirical evaluations demonstrate the practical effectiveness.

We used the SYNAPSE-G technique in a real-world project that aimed to evaluate content against a substantial number of abuse policies. Given the quantity and specificity of these policies, labeled data for initial training was unavailable or extremely rare. We employed an LLM to produce synthetic examples for various abuse policies to run positive mining using SYNAPSE-G. In our real-world evaluations, Label Propagation (LP) was chosen over the Bipartite approach due to its ability to construct large real-world data graphs once and iteratively reuse them with additional synthetic datasets across all policies, eliminating the need for repeated or complicated incremental graph reconstruction. In the present paper, we showed that LP indeed outperforms the Bipartite approach and validate this practical framework for improving rare event detection across various real-world contexts.


\bibliography{references}

\clearpage
\appendix
\onecolumn
\section{Rare Event Designation for MHS} \label{apx:mhs-data}

For labeling, the MHS dataset is more complex and relies on further preprocessing to define a specific rare event. 
Specifically, the dataset comprises social media comments (YouTube, Reddit, Twitter) annotated across 10 ordinal labels to derive a continuous ``hate speech score'', with each sentence also being labeled according to the specific groups or demographics targeted. 
As such, we here define a specific rare event binary label for MHS: hate speech targeting transgender individuals. 
Concretely, we create a label which is positive if any annotator marked a comment as targeting any of the sub-categories: transgender men, transgender women, or unspecified transgender individuals.
We do this by applying a logical OR across the three noted subcategory annotations. 
This resulted in 2,598 comments (6.5\%) being labeled positive, representing an organically imbalanced dataset relevant to real-world challenges.

\section{Algorithm Pseudocode} \label{apx:pseudo}
\begin{algorithm}[h]
\caption{Iterative Bipartite Graph (IBG)}
\label{alg:ibg}
\begin{algorithmic}[1]
\REQUIRE Unlabeled data $\mathcal{D}_U$, positive seeds $\mathcal{D}_S$, threshold $\tau$, max degree $d_{max}$, iterations $T$
\ENSURE Labeled data $\mathcal{L}$

\STATE Initialize $V_P = \mathcal{D}_S$, $\mathcal{L} = \mathcal{D}_S$
\FOR{$t = 1$ to $T$}
  \STATE $V_U = \mathcal{D}_U$  \hfill // Remaining unlabeled data
  \STATE Construct bipartite $G_B = (V_P, V_U, E_B)$
    \FOR{$v_i \in V_P$}
        \FOR{$v_j \in V_U$}
          \IF{$\texttt{cos\_sim}(v_i, v_j) > \tau$}
                \STATE $E_B \leftarrow E_B \cup e_{ij} $
          \ENDIF
        \ENDFOR
          \STATE Sort $N(v_i)$ in $V_U$ by \texttt{cos\_sim} (desc.)
        \STATE Keep top $d_{max}$ neighbors, remove other edges from $E_B$
    \ENDFOR
    \STATE $\mathcal{B}_t \leftarrow v \in V_U$ connected to $V_P$ in $G_B$
    \STATE Query labels $\mathcal{L}_t$ for $\mathcal{B}_t$
    \STATE $V_P \leftarrow V_P \cup \{v \in \mathcal{B}_t \mid \text{label}(v) = \text{positive}\}$
    \STATE $\mathcal{L} \leftarrow \mathcal{L} \cup \mathcal{L}_t$
    \STATE $\mathcal{D}_U \leftarrow \mathcal{D}_U \setminus \mathcal{B}_t$
\ENDFOR
\RETURN $\mathcal{L}$
\end{algorithmic}
\end{algorithm}

\begin{algorithm}[h]
\caption{Label Propagation}
\label{alg:label_prop}
\begin{algorithmic}[1]
\REQUIRE Similarity graph $G=(V,E)$, initial (partial) labels $Y_0$, iterations $T$
\ENSURE  Final label assignments $Y_T$

\STATE Initialize $Y^{(0)} = Y_0$
\FOR{$t = 1$ to $T$}
    \STATE Construct normalized adjacency matrix:
    \[ W_{ij} = \begin{cases}
        \frac{1}{\text{deg}(v_i)} & \text{if } (v_i, v_j) \in E \\
        0 & \text{otherwise}
    \end{cases} \]
    \STATE $Y^{(t)} = W Y^{(t-1)}$
    \STATE Set $Y_i^{(t)} = Y_{0,i}$ for all $i$ with labels in $Y_0$
\ENDFOR
\RETURN $Y^{(T)}$.

\end{algorithmic}
\end{algorithm}

\section{Omitted Proofs} \label{apx:proofs}
\subsection{Proof of Proposition~\ref{prop:precision-recall}}

\begin{proof}
$Q = S \cup N(S_+) = S_+ \cup S_- \cup N(S_+) = S_- \cup N(S_+)$. Since $S$ is an independent set (Assumption \ref{assump:diversity}), $S_- \cup N(S_+)$ is disjoint. Thus,
\begin{align*}
|Q| &= |S_-| + |N(S_+)| \\
&= (1-p)|S| + h(S_+)p|S| \\
&= (1 - p + ph(S_+))|S|.
\end{align*}
Let $S_1 \subseteq N(S_+) \setminus S_+$ be vertices in $N(S_+) \setminus S_+$ adjacent to exactly one vertex in $S_+$, and $S_2$ be those adjacent to exactly two. Let $P_1$, $P_2$ be the number of positive examples in $S_1$, $S_2$, respectively.
\begin{align*}
\mathbb{E}[P \mid S] &= \mathbb{E}[|S_+| + P_1 + P_2 \mid S] \\
&= |S_+| + q_1|S_1| + q_2|S_2|.
\end{align*}
We have:
\begin{align}
|S_+| + |S_1| + |S_2| &= |N(S_+)| \label{eq:1}\\
d|S_+| + |S_+| - |S_2| &= |N(S_+)| \label{eq:2}\\
|N(S_+)| &= |S_+|h(S_+). \label{eq:3}
\end{align}
Equation \eqref{eq:1} counts vertices in $N(S_+)$. Equation \eqref{eq:2} counts edges between $S_+$ and $N(S_+)$, subtracting $|S_2|$ once (as each is counted twice). Equation \eqref{eq:3} is from the definition of $h(S_+)$. Solving \eqref{eq:1}-\eqref{eq:3}:

\begin{align*}
|S_1| &= (2h(S_+) - d - 2)|S_+| \\
|S_2| &= (d + 1 - h(S_+))|S_+| \\
\mathbb{E}[P \mid S] &= |S_+| (1  + q_1(2h(S_+) - d - 2) \\
&\quad+ q_2(d + 1 - h(S_+))) \\
&= p|S| ((1 - 2q_1 + q_2) \\
&\quad+ (q_2 - q_1)d + (2q_1 - q_2)h(S_+)).
\end{align*}
Therefore,
\begin{align*}
&\mathbb{E}[\mathrm{Precision} \mid S] \\
&= \frac{p|S|((1 - 2q_1 + q_2) + (q_2 - q_1)d + (2q_1 - q_2)h(S_+))}{(1 - p + ph(S_+))|S|} \\
&= (2q_1 - q_2) + \frac{1 + q_2\left(d + \frac{1}{p}\right) - q_1\left(d + \frac{2}{p}\right)}{\frac{1-p}{p} + h(S_+)}.
\end{align*}
If $1 + q_2\left(d + \frac{1}{p}\right) - q_1\left(d + \frac{2}{p}\right) > 0$ (i.e., $p > \frac{2q_1 - q_2}{1 + (q_2 - q_1)d}$), $\mathbb{E}[\mathrm{Precision} \mid S]$ decreases with $h(S_+)$.
Now, we compute the derivative of $\mathbb{E}[\mathrm{Precision} \mid S]$ with respect to $p$:
\begin{align*}
&\frac{\partial}{\partial p} \mathbb{E}[\mathrm{Precision} \mid S] \\
&= \frac{(1 - q_1(2+d) + q_2(1+d)) + h(S_+)(2q_1 - q_2)}{(1 - p + ph(S_+))^2} \\
&= \frac{1 + d(q_2 - q_1) + (h(S_+) - 1)(2q_1 - q_2)}{(1 - p + ph(S_+))^2}.
\end{align*}
Since $h(S_+) \leq d + 1$, $d \geq h(S_+) - 1$. Thus,
\begin{align*}
&\frac{\partial}{\partial p} \mathbb{E}[\mathrm{Precision} \mid S] \\
&\geq \frac{1 + (h(S_+) - 1)(q_2 - q_1) + (h(S_+) - 1)(2q_1 - q_2)}{(1 - p + ph(S_+))^2} \\
&= \frac{1 + (h(S_+) - 1)q_1}{(1 - p + ph(S_+))^2} > 0.
\end{align*}
Therefore, $\mathbb{E}[\mathrm{Precision} \mid S]$ is strictly increasing with respect to $p$.

Finally,
\begin{align*}
\mathbb{E}[\mathrm{Recall} \mid S] &= \frac{p|S|}{ |V|} \bigl((1 - 2q_1 + q_2) \\
&\quad+ (q_2 - q_1)d + (2q_1 - q_2)h(S_+)\bigr).
\end{align*}
Since $q_2 < 2q_1$, $\mathbb{E}[\mathrm{Recall} \mid S]$ increases with $p$ and $h(S_+)$.
\end{proof}

 
\end{document}